\documentclass{llncs}
\usepackage{times}
\usepackage{graphicx}
\usepackage{latexsym}

\begin{document}

\title{From cells to islands: An unified model of \\ cellular parallel Genetic Algorithms}
\author{David Simoncini, Philippe Collard, S\'ebastien Verel, Manuel Clergue}
\institute{Universit\'e Nice Sophia-Antipolis/CNRS \\
 \email{$\lbrace$simoncin,pc,verel,clerguem$\rbrace$@i3s.unice.fr}}
\maketitle

\begin{abstract}

This paper presents the Anisotropic selection scheme for cellular Gen\-etic
Algorithms (cGA). This new scheme allows to enhance diversity and to control
the selective pressure which are two important issues in Genetic Algori\-thms,
especially when trying to solve difficult optimization problems. Varying the
aniso\-tropic degree of selection allows swapping from a cellular to an island
model of parallel genetic algorithm. Measures of performances and diversity
have been performed on one well-known problem: the Quadratic Assignment Problem which is
known to be difficult to optimize. Experiences show that, tuning the
anisotropic degree, we can find the accurate trade-off between cGA and island
models to optimize performances of parallel evolutionary algorithms. This
trade-off can be interpreted as the suitable degree of migration among
subpopulations in a parallel Genetic Algorithm.

\end{abstract}

\section*{Introduction}
In the context of cellular genetic algorithm (cGA), this paper proposes the
Anisotropic selection as a new selection scheme which accurately allows to
adjust the selective pressure and to control the
 exploration/exploitation ratio. This new class of evolutionary algorithms is
supervised in a continuous way by an unique real parameter $\alpha$ in the
range [-1..1]. The work described in this paper is an attempt to provide
a unified model of parallel genetic algorithms (pGA) from fine grain massively parallel GA 
(cGA) to coarse grain
parallel model (island GA). As extreme cases, there are the cGA that assumes one
individual resides at each cell, and at the opposite, a pGA where
distinct subpopulations execute a standard GA; between them we find models of pGA
where migration allows to exchange to some extend genetic information between
subpo\-pulations. Thus the search dynamics of our family of pGA can vary from 
a diffusion to a migration process. To illustrate our approach we used one well-known problem:
the Quadratic Assignment Problem (QAP). We study the performances of our class of
parallel evolution\-ary algorithms on this problem and we show that there is a threshold for parameter $\alpha$ according to the average performances. Section 1 gives a description of the cGA and 
the island models. Section 2 introduces the anisotropic parallel Genetic Algorithms (apGA) and 
the anisotropic selection scheme. Section 3 is a presentation of the test problem: the QAP, and gives the performances of the apGA on the QAP. Finally, a study on population 
genotypic diversity is made in section 4. 
 
\section{Background}

This section introduces the concepts of Cellular and Island Models of paral\-lel genetic algorithms.

\subsection{Cellular Genetic Algorithms}

The Cellular Genetic Algorithms are a subclass of Evolutionnary Algorithms in which the population is generally embedded on a two dimensional toroidal grid.
In this kind of algorithms, exploration and population diversity are enhanced thanks to the existence of small overlapped neighborhoods \cite{SpiessensM91}. An 
 individual of the population is placed on each cell of the grid and represents a solution of the problem to solve.
 An evolutionnary process runs simultaneously on each cell of the grid, selecting parents from the neighborhood of the cells and applying operators 
 for recombination, mutations and replacement for further generations. 
 Such a kind of algorithms is especially well suited for complex problems \cite{JongS95}.
One of the interests of cGA is to slow down the convergence of the population among a single individual. Complex problems often have many 
local optima, so if the best individual spreads too fast in the population it will improve the chances to reach a local optimum of the search 
space. Slowing down the convergence speed can be done by slowing down the selective pressure on the population.  

\subsection{Island Model of pGA}
Cellular genetic algorithms and Island Model genetic algorithms are two kinds of Parallel genetic algorithms. The first one is a \textit{fine grain} massively parallel implementation that assumes 
one individual resides at each cell. The second one, using distinct subpopulations, is a 
\textit{coarse grain} parallel model; Each subpopulation executes as a standard genetic 
algorithm, and occasionally the subpopulations would exchange a few strings: \textit{migration} 
allows subpopulations to share genetical material \cite{Gorges91}. Many topolo\-gies can be defined to connect the islands. In the basic island model, migration can occur between any subpopulations, whereas in the 
\textit{stepping stone} model islands are disposed on a ring and migration is restricted to 
neighboring islands.

\section{Anisotropic Parallel Genetic Algorithms}

This section presents the \textit{anisotropic parallel Genetic Algorithms},
which is a family of parallel genetic algorithms based on cellular GA in
which anisotropic selection is used.

\subsection{Definition}

The Anisotropic selection is a selection method in which the neighbors of a cell may have different probabilities to be selected.
The Von Neumann neighborhood of a cell $C$ is defined as the sphere of radius $1$ centered at $C$ in manhattan distance. 
 The Anisotropic selection assigns different 
 probabilities to be selected to the cells of the Von Neumann neighborhood according to their position. The probability to choose the center cell $C$ 
remains fixed at $\frac15$. Let us call $p_{ns}$ the probability of choosing the cells North ($N$) or South ($S$) and $p_{ew}$ the probability of choosing the cells East  
($E$) or West ($W$). Let $\alpha \in [-1;1]$ be the control parameter that will determine the probabilities $p_{ns}$ and $p_{ew}$. This parameter will be called 
the \textit{anisotropic degree}.
The probabilities $p_{ns}$ and $p_{ew}$ can be described as: \\
$$p_{ns}=\frac{(1-p_c)}{2}(1+\alpha)$$
$$p_{ew}=\frac{(1-p_c)}{2}(1-\alpha)$$ 
Thus, when $\alpha=-1$ we have $p_{ew}=1-p_c$ and $p_{ns}=0$. When $\alpha=0$, we have $p_{ns}=p_{ew}$ and when $\alpha=1$, we have $p_{ns}=1-p_c$ and $p_{ew}=0$.

\begin{figure}[ht!]
\begin{center}
\includegraphics[width=3cm,height=3cm]{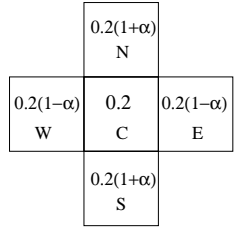} 
\caption{Von Neumann neighborhood with probabilities to choose each neighbor}
\end{center}
\label{VN}
\end{figure}

Figure $1$ shows a Von Neumann Neighborhood with the probabilities to select each cell as a function of 
$\alpha$.

The Anisotropic Selection operator works as follows. 
For each cell it selects $k$ individuals in its neighborhood ($k \in [1;5]$). The $k$ individuals participate to a tournament and the winner replaces 
the old individual if it has a better fitness or with probability $0.5$ if the fitnesses are equal. 
When $\alpha=0$, the anisotropic selection is equivalent to a standard tournament selection and when $\alpha=1$ or $\alpha=-1$ the anisotropy is maximal and we 
have an uni-dimensional neighborhood with three neighbors only. In the following, considering the grid symmetry we will consider $\alpha \in [0;1]$ only: when $\alpha$ is in the 
range [-1;0] making a rotation of $90^{\circ}$ of the grid is equivalent to 
considering $\alpha$ in the range [0;1]. 
When the anisotropic degree is null,
there is no anisotropy in selection method,
the apGA corresponds to the standard cellular GA.
When the anisotropic degree is maximal,
 selection is computed between individuals in the same column only,
the apGA is then an island model where
each subpopulation is a column of the grid structured as a ring of cells with no interactions 
between subpopula\-tions.
When the anisotropic degree is set between low and maximum value,
according to selection,
a number of individuals can be copied from one subpopulation (i.e. column) to the adjacent columns.
Thus the anisotropic degree allows to define a family of parallel GA
from a cellular model to an island model.

In standard island model,
the migration rate is defined as the number of individuals
which are swap between subpopulations
and migration intervals is the frequency of migration.
In apGA,
the migration process is structured by the grid.
Only one parameter (the anisotropic degree) is needed to tune the migration policy. There is a difference between migration in a standard island model and migration in an apGA. In an apGA it can only happen (when the anisotropic degree allows it) between nearest neighbors 
in adjacent columns. Migration in that latter case is diffusion as it happens 
 in the standard cGA model, except that the direction is controllable.  
In the following sections,
we study the influence of this parameter on selection pressure,
performances and population diversity.

\subsection{Takeover times and apGAs}
\label{section3}

\begin{figure}[ht!]
\begin{center}
\includegraphics[width=6cm,height=6cm]{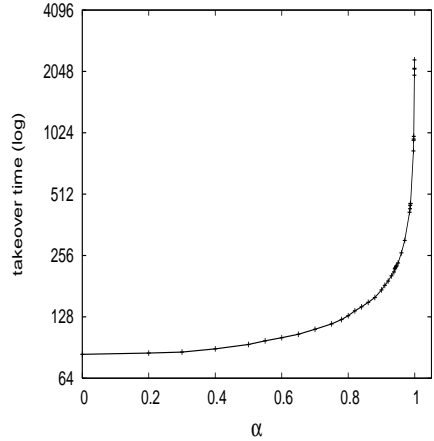}
\end{center}
\caption{Average of the takeover time as a function of the anisotropic degree $\alpha$.}
\label{fig-tak-fuzzy}
\end{figure}

The selective pressure is related to the population diversity in cellular genetic algorithms. One would like to slow down the selective pressure when trying to 
solve multimodal problems in order to prevent the algorithm from converging too fast upon a local opti\-mum. On the opposite side, when there is no danger 
of converging upon a local optimum, one would like to increase the selective pressure in order to obtain a good solution as fast as possible. 
A common analytical approach to measure the selective pressure is the computation of the takeover time \cite{Rudolph} \cite{Sprave}. It is the number of generations needed for the best individual to conquer the whole
 grid when the only active operator is the selection \cite{GoldbergD90}. Figure \ref{fig-tak-fuzzy} shows the influence of the anisotropic degree on the takeover time.
This figure represents the average takeover times observed on $1000$ runs on a $32 \times 32$ grid for different anisotropic degrees.
 It shows that the selective pressure is decreasing while increasing anisotropy. These results confirm that the anisotropic selection gives to the algorithm
 the ability to control accurately the selective pressure.     
They are fairly consistent with our expectation that selection intensity decreases when the anisotropic degree increases. However, the correlation between takeover and anisotropy is not linear; it fast increases after the value $\alpha=0.9$.

\section{Test problem}

This section presents tests on one well-known instance of the Quadratic Assignment Problem which is known to be difficult to optimize. Our aim is to study the dynamics of the apGA for 
different tunings, and not to obtain better performances than other optimization techniques. 
Still, the apGA is implicitely compared to a cellular genetic algorithm when the anisotropic
degree is null ($\alpha=0$).

\subsection{The Quadratic Assignment Problem}
\label{section6}
We experimented the family of apGAs on a Quadratic Assignment Problem (QAP): Nug30.
 Our purpose here is not to obtain better results with respect to other optimization methods, but rather to observe the behavior of
apGAs. Especially we go in the search of a threshold for the anisotropic degree.

The QAP is an important problem in theory and practice as well. It was introduced by Koopmans and Beckmann
in 1957 and is a model for many practical problems \cite{Koopmans57}. 
The QAP can be described as the problem of assigning a set of facilities to
a set of locations with given distances between the locations and given flows between the 
facilities. The goal is to place the facilities on locations in such a way that the sum
 of the products between flows and distances is minimal.
\linebreak
Given $n$ facilities and $n$ locations, two $n \times n$ matrices $D=[d_{ij}]$ and $F=[f_{kl}]$
 where $d_{ij}$ is the distance between locations $i$ and $j$ and $f_{kl}$ the flow between 
 facilities $k$ and $l$, the objective function is: \\
\begin{displaymath}
\Phi = \sum_{i}\sum_{j}d_{p(i)p(j)}f_{ij}
\end{displaymath}
where $p(i)$ gives the location of facility $i$ in the current permutation $p$.
\linebreak
 Nugent, Vollman and Ruml proposed a set of problem instances of different sizes noted for their difficulty \cite{Nugent68}. The instances they proposed are known to have multiple local optima, so they are difficult for a genetic algorithm. We experiment our algorithm on the 30 variables instance called Nug30.



\subsection{Setup}

We use a population of 400 individuals placed on a square grid ($20\times 20$). Each individual represents
 a permutation of $\lbrace 1,2,...,30 \rbrace$. We need a special crossover that preserves the permutations:
\begin{itemize}
\item
  Select two individuals $p_1$ and $p_2$ as genitors.
\item
  Choose a random position $i$.
\item
  Find $j$ and $k$ so that $p_1(i) = p_2(j)$ and $p_2(i) = p_1(k)$.
\item
  exchange positions $i$ and $j$ from $p_1$ and positions $i$ and $k$ from $p_2$.
\item
  repeat $n/3$ times this procedure where $n$ is the length of an individual.
\end{itemize}

This crossover is an extended version of the UPMX crossover proposed in \cite{Migkikh}.
The mutation operator consist in randomly selecting two positions from the individual
 and exchanging those positions. The crossover rate is 1 and we do a mutation per individual.
We perform 500 runs for each anisotropic degree. Each run stops after 1500 
generations.

\subsection{Experimental results}

\begin{figure}[ht!]
\begin{center}
\includegraphics[width=6cm,height=6cm]{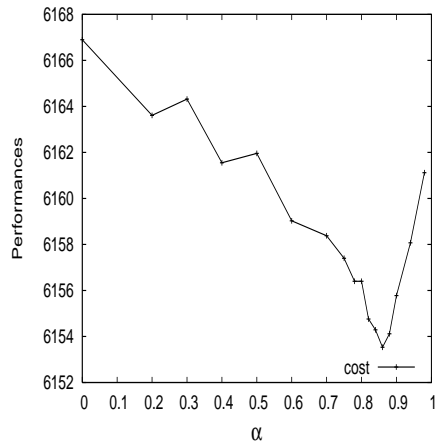}
\end{center}
\caption{Average costs as a function of $\alpha$ for the QAP.}
\label{alphaperf}
\end{figure}

Figure \ref{alphaperf} shows the average performance of the algorithm towards $\alpha$ on the QAP: 
for each value of $\alpha$ we average the best solution of each run.
The purpose here is to minimize the fitness function values.
 The performances are growing with $\alpha$ and then fall down as $\alpha$ is getting closer to its limit value.
The best average performance is achieved for $\alpha=0.86$. This threshold probably corresponds to a good exploration/exploitation trade-off: the algorithm favors propagation of good solutions in the vertical direction with few interactions on the left or the right sides. This kind of dynamics is well adapted to this multi-modal problem as we can reach 
local optima on each columns of the grid and then migrate them horizontally to find new solutions. 
 The worst average performance 
is observed for $\alpha=0$ when the apGA is a cellular GA. $\alpha=0.86$ corresponds to the optimal trade-off 
between cellular and island models for this problem, with the best migration rate between subpopulations. In our model, the migration rate is not the number of individuals
which are swap between subpopulations, but the probabili\-ty for the selection operator 
to choose two individuals from separate columns: two indivi\-duals from separate subpopulations 
would then share information.
 We can tell that there is an optimal migration rate that is induced by 
the value of the anisotropic degree $\alpha$. 
Performances would probably improve if the migration rate did not stay static during the search process. As in \cite{Alba05}, we can define some criteria to self-adjust the anisotropic degree along generations.

\section{Diversity in apGAs}

To understand better why we observe influence of the anisotropic parameter
on perform\-ances, we felt it is important to measure genetic diversity during
runs. We studied changes in diversity during runs according to the whole
grid, the rows and the columns. 

This section presents measures on population diversity in an apGA for the QAP. We conducted experiences on the 
average population diversity observed along generations on $100$ independent runs for each anisotropic degree.
We made three measures on the population diversity. First, we computed the global population diversity $gD$: 
$$
gD=(\frac{1}{\sharp r \sharp c})^2\sum_{r_1,r2}\sum_{c_1,c_2} d(x_{r_1c_1},x_{r_2c_2})
$$ 
where $d(x_1,x_2)$ is the distance between individuals $x_1$ and $x_2$.
The distance used is inspired from the Hamming distance: It is the number of locations that
differs between two individuals divided by their length $n$.

Then, we made measures on diversity inside subpopulations (vertical diversity)
 and diversity between subpopulations (horizontal diversity). The vertical (resp. horizontal) diversity is the sum
of the average distance between all individuals in the same column (resp. row) divided by the 
number of columns (resp. rows):

$$
vD=\frac{1}{\sharp r}\frac{1}{\sharp c^2} \sum_{r}\sum_{c_1,c_2} d(x_{rc_1},x_{rc_2})
$$ 

$$
hD=\frac{1}{\sharp c}\frac{1}{\sharp r^2}\sum_{c}\sum_{r_1,r_2} d(x_{r_1c},x_{r_2c})
$$ 
where $\sharp r$ and $\sharp c$ are the number of rows and columns in the grid.

\begin{figure}[ht!]
\begin{center}
\begin{tabular}{cc}
\includegraphics[width=5cm,height=5cm]{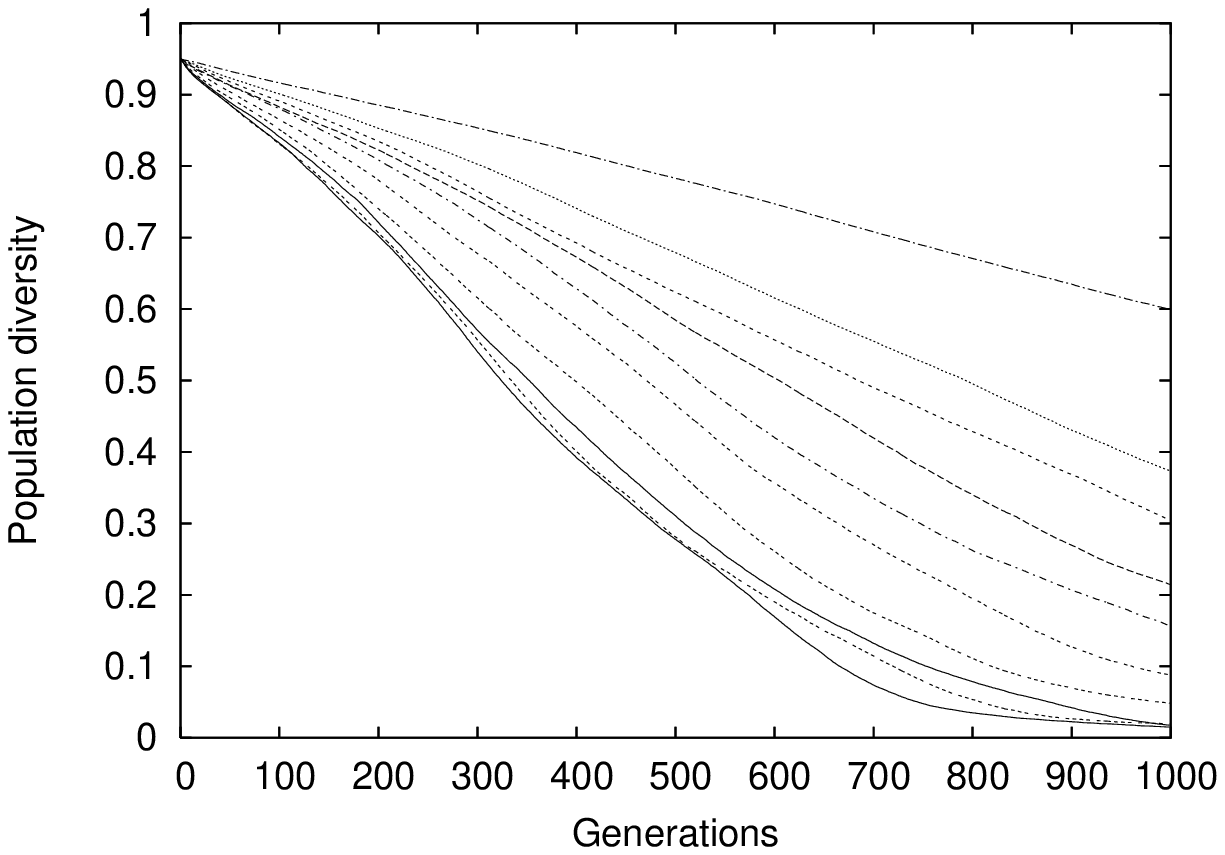} &
\includegraphics[width=5cm,height=5cm]{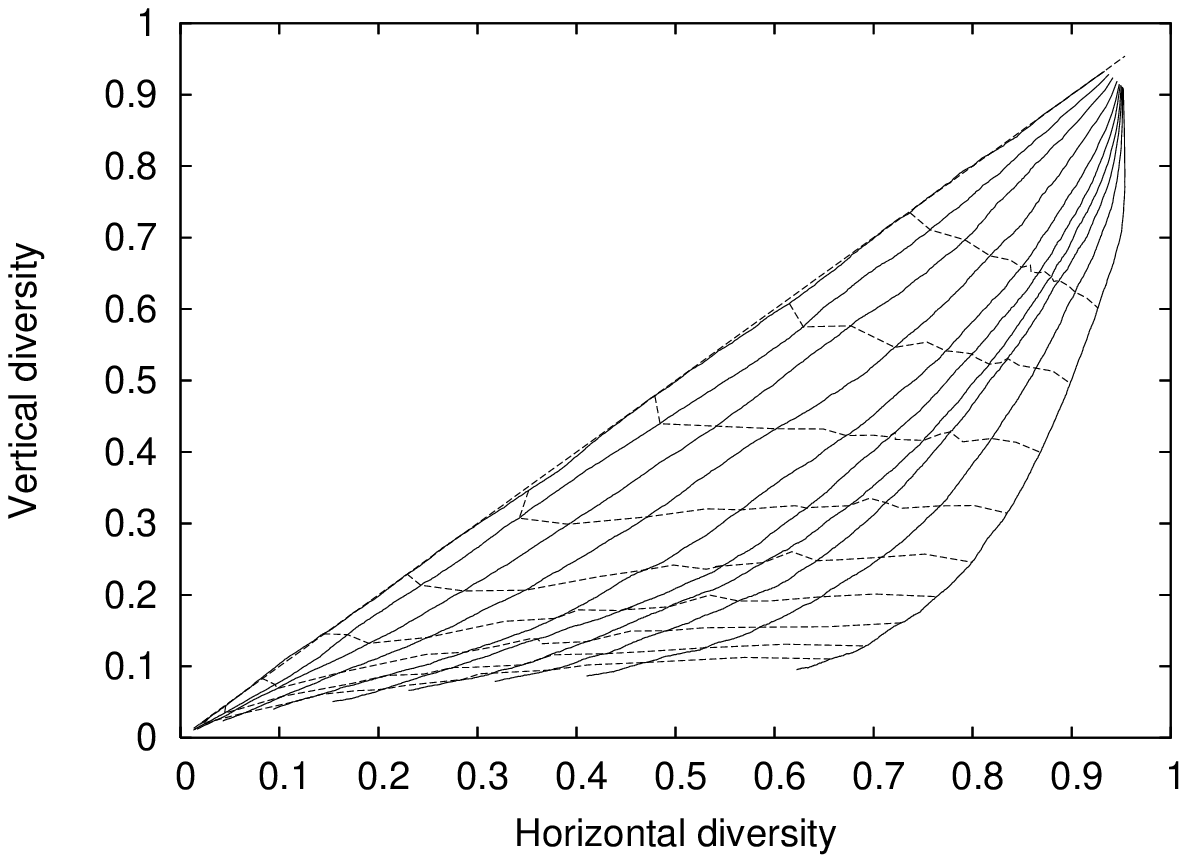} \\
 (a) &
 (b) \\
\end{tabular}
\end{center}
\caption{Global population diversity against generation, with increasing $\alpha$ from bottom to top (a) and vertical diversity against horizontal diversity, with increasing $\alpha$ from left to right (b).}
\label{globdiv}
\end{figure}
  

\begin{figure}[ht!]
\begin{center}
\fbox{\begin{tabular}{cccccc}
    &
$1$ &
$200$ &
$500$ & 
$1000$ & 
$2000$ \\
$0$ &
\includegraphics[width=2cm,height=2cm]{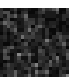} &
\includegraphics[width=2cm,height=2cm]{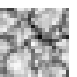} &
\includegraphics[width=2cm,height=2cm]{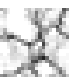} &
\includegraphics[width=2cm,height=2cm]{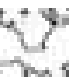} &
\includegraphics[width=2cm,height=2cm]{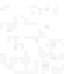} \\
$0.5$ &
\includegraphics[width=2cm,height=2cm]{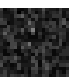} &
\includegraphics[width=2cm,height=2cm]{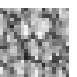} &
\includegraphics[width=2cm,height=2cm]{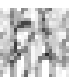} &
\includegraphics[width=2cm,height=2cm]{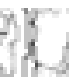} &
\includegraphics[width=2cm,height=2cm]{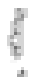} \\
$0.7$ &
\includegraphics[width=2cm,height=2cm]{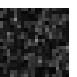} &
\includegraphics[width=2cm,height=2cm]{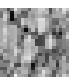} &
\includegraphics[width=2cm,height=2cm]{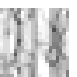} &
\includegraphics[width=2cm,height=2cm]{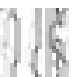} &
\includegraphics[width=2cm,height=2cm]{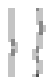} \\
$0.86$ &
\includegraphics[width=2cm,height=2cm]{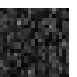} &
\includegraphics[width=2cm,height=2cm]{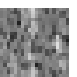} &
\includegraphics[width=2cm,height=2cm]{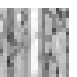} &
\includegraphics[width=2cm,height=2cm]{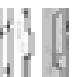} &
\includegraphics[width=2cm,height=2cm]{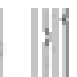} \\
$0.98$ &
\includegraphics[width=2cm,height=2cm]{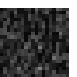} &
\includegraphics[width=2cm,height=2cm]{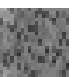} &
\includegraphics[width=2cm,height=2cm]{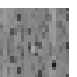} &
\includegraphics[width=2cm,height=2cm]{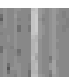} &
\includegraphics[width=2cm,height=2cm]{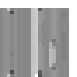} \\
\end{tabular}}
\end{center}
\caption{Local diversity in the population along generations (left to right) for increasing 
$\alpha$ (top to bottom) }
\label{divers}
\end{figure}

Figure \ref{globdiv}(a) shows the average global diversity observed on the $1000$ first generations during $100$ runs on the QAP.
The curves from bottom to top correspond to increasing values
 of $\alpha$ from zero to nearly one.
Experiments measuring genetic diversity show that small migration rate ($\alpha$ close to one) causes islands 
to dominate others and retain global diversity without being able to exchange solutions to produce better results.
At the opposite, for the cellular model, as $\alpha$ is closed to zero, global diversity falls near to zero 
after $800$ generations causing premature convergence and negatively affects performances (see figure \ref{alphaperf}).
Analysis on the QAP show the necessity of maintaining diversity to produce new results and the necessity to have enough information exchanges between columns. 

Figure \ref{globdiv}(b) represents the vertical diversity against the horizontal diversity. The contour lines plotted
every $100$ generations give some information on the speed of decrease of diversity. The more the migration rate decreases (i.e. $\alpha$ increases), the more 
the diversity is maintained on each row and subpopulations converge in each column. The vertical and horizontal diversities are decreasing with the same speed for the cellular model ($\alpha=0$) and lower number of interactions between subpopulations helps the 
algorithm to maintain diversity on the rows when $\alpha$ is high.

Figure \ref{divers} shows snapshots of the population diversity during one single run at diffe\-rent generations. The snapshots are taken from left to right at generations $1$, $200$, $500$, $1000$ and $2000$. The parameter $\alpha$ takes values in $\lbrace 0,0.5,0.7,0.86,0.98 \rbrace$ from top 
to bottom.
 Each snapshot shows the genotypic diversity in the neighborhoods of all cells on the grid.
 Color black means maximum diversity and color white means that there is no more diversity in the cell's neighborhood.
Those snapshots help to understand the influence of the anisotropic selection on the genotypic diversity.
First, we can see that the anisotropic degree influences the 
dynamic of propagation of good individuals on the grid. This propagation is the cause 
of the loss of diversity in the population. In the standard cellular model ($\alpha=0$),
 good individuals propagate roughly circularly. If we slightly privilege the vertical direction
 ($\alpha=0.5$) the circles become elliptical. As $\alpha$ increases, the dynamic changes
 and good individuals propagate column by column. For extreme values of the anisotropic degree
 ($\alpha$ close to $1$) the migration rate is so low that good individuals are stuck in 
the subpopulations and the sharing of genetic information with other subpopulations
 is seldom observed. 
In that case, the selective pres\-sure is too low and it negatively affects performances.
The crossover operator doesn't have any effect in the white zones, since they represent 
cells with no more diversity in their neighborhoods. For the standard cellular case, 
interactions between cells may have some effects on performances only at the frontier 
between the circles. It represents a little proportion of cells on the grid after a
 thousand generations. For $\alpha=0.86$, we can see vertical lines of diversity, which 
means that good individuals appear in each subpopulations. For example, when we see 
two adjacent columns colored in grey it means that those columns have been 
colonized by two different individuals. At generation $2000$,
 a good individual has colonized the left of the grid but he still can share information
 with individuals in the grey zones. This means that the migration rate between 
subpopula\-tions is strong enough to guarantee the propagation of the genetic information 
through the whole grid.   
This study showed that the dynamic of the propaga\-tion of individuals on the grid is 
strongly related to the anisotropic degree. Once again, it would be interesting to see 
what kind of dynamic appears if we define a local criteria to auto-adapt $\alpha$ during 
a run. This parallel model of GA allows to tune separately the anisotropic degree for 
each cell on the grid and measures during the search process can help to adjust locally 
the selective pressure. 
\clearpage

\section*{Conclusion and Perspectives}
This paper presents a unified model of parallel Genetic Algorithms where granularity can
be continuously tuned from fine grain to coarse grain parallel model. This
family is based on the new concept of anisotropic selection. We analysed the dynamics of this class of 
pGAs on the well-known QAP
problem. We have shown that the anisotropic degree plays 
a major role with regard to the average fitness found. Perform\-ances of the apGA increases 
with $\alpha$ until a threshold value ($\alpha=0.86$). After this threshold, the migration rate between subpopulations in columns may be too small to generate good solutions. A study on 
local diversity shows the interactions between cells for different tunings of 
the apGA.
The dynamic of propagation of individuals, which is strongly related to 
the genotypic diversity in the population, is dependent from the anisotropic degree of the apGA.
 Propagation of good individuals is done in circles for low 
values of $\alpha$ and turns to vertical lines for high values of $\alpha$.
Diversity is maintained in the population when the anisotropic degree is high, but when it reaches values close to the extreme case the few interactions between columns penalize the performances
 of the algorithm.   
These experimental results lead us to suggest to adjust dynamically the
migration ratio during a run:
  by tuning the control parameter
$\alpha$, it would be possible to make the algorithm to self-adjust the migration level,
 depending on global or local measures. While theorical and experimental studies on island models
 are difficult due to their complexity, the apGA model could be used as a simple framework for 
calculations on parallel GA.
Naturally it would be worth seeing how properties described in 
this paper extend for even more complex problems.

\bibliographystyle{abbrv}

\end{document}